# Technical Report: Automated Optical Inspection of Surgical Instruments

Zunaira Shafqat*, Atif Aftab Ahmed Jilani (*Member, IEEE*), and Qurrat Ul Ain

*Corresponding Author: zunaira.shafqat@isb.nu.edu.pk
Contributing authors: atif.jilani@nu.edu.pk ; quratul.ain.v@isb.nu.edu.pk

**Abstract —** **In the dynamic landscape of healthcare, the significance of maintaining the highest standards in surgical instruments cannot be overstated. This report explores the diverse realm of surgical instruments and their associated defects, emphasizing their pivotal role in ensuring the success of surgical procedures. With potentially fatal consequences for any defects, precision in manufacturing becomes paramount.This study focuses on the Pakistani surgical instruments manufacturing industry, a significant contributor to the country's exports. Recognizing its importance, the report sheds light on the identification and rectification of defects within these instruments. Not only does this scrutiny prevent potential losses for manufacturers but, more crucially, it safeguards patients' lives. The collaboration with industry leaders Daddy D Pro and Dr. Frigz International, renowned trailblazers in the Sialkot surgical industry, provides invaluable insights into the analysis of defects in Pakistani surgical instruments. This partnership signifies a commitment to advancing defect detection methodologies, thereby elevating the overall quality standards in the manufacturing process. The scope of this report is to comprehensively identify various surgical instruments manufactured in Pakistan and analyze their associated manufacturing defects. By focusing on quality assurance, the document aims to serve as a valuable resource for manufacturers, healthcare professionals, and regulatory bodies. The insights gained from this exploration contribute to the enhancement of surgical instrument standards, ensuring a safer and more reliable healthcare environment. The collaborative efforts of DaddyDPro and Dr. Frigz as industry partners underscore a groundbreaking project aimed at revolutionizing defect detection in Pakistani surgical instruments. This initiative, combining industry expertise and cutting-edge technology, represents a paradigm shift in elevating the quality standards of surgical instruments in the country.**

## I. INTRODUCTION

As the healthcare landscape continues to evolve, the importance of ensuring the highest standards in surgical instruments cannot be overstated. Ensuring precision of surgical instruments is crucial for success of surgical procedures and any defects in them, could potentially have, fatal consequences

[14]. Daddy D Pro and Dr. Frigz International, recognized as trailblazers in the Sialkot surgical industry, have provided their valuable insights and analysis of defects in surgical instruments produced in Pakistan. This partnership signifies a commitment to advancing defect detection methodologies and elevating the overall quality standards in the manufacturing process. Surgical instruments are a vital part of modern surgery, and the global market for these tools is worth over $30 billion [1]. Pakistan plays a significant role in this industry, exporting surgical equipment valued at around $359 million annually. According to the Pakistan Surgical Instruments Manufacturers Association (PSIMA) [38], Pakistani companies produce more than 170 million medical instruments each year. These instruments fall into two categories: disposable (60% of exports) and reusable (40% of exports).

The surgical instrument industry faces a major challenge: ensuring consistent quality. This is especially important given the increasing competition in the global market and stricter quality regulations imposed by developed countries [2]. While Pakistan exports a significant portion of its surgical instrument production (around 95%), quality issues can hinder further growth. Many Pakistani manufacturers subcontract the initial stages of instrument production to smaller workshops or even workers' homes. The finishing and quality checks are then carried out in-house before export, often against European Union or US standards. However, relying on manual visual inspection can be inconsistent, leading to entire batches of instruments being rejected and export orders being canceled [3]. Manual inspection, the traditional method for identifying defects in surgical instruments, is labor-intensive, prone to human error, and often inconsistent. This poses a significant risk in high-stakes environments like operating rooms, where faulty instruments could lead to severe consequences. The inadequate inspection of sterilized instruments can lead to compromised sterility, which underscores the need for more reliable, automated methods of defect detection and classification [5]. Recent advancement in deep learning and computer vision have paved the way for automated solutions that can address the limitations of manual inspection. In particular, image-based defect detection methods offer a non-invasive and scalable approach to identifying defects in surgical instruments. Studies have explored the use of deep learning models for instrument classification and defect detection in real-time scenarios. For example, Campos-Montes et al. [26] implemented a computer-assisted surgical instrument system using deep learning, achieving high accuracy in identifying

This work is supported by the National Research Program for Universities (NRPU), Higher Education Commission (HEC) Pakistan, under Project No. 15872, titled "Automated Optical Inspection for Industrial Surgical Instruments."

Authors are with National University of Computer and Emerging Sciences (FAST-NUCES), Islamabad, Pakistan

both instrument types and defects. However, many existing studies focus on general cleanliness or instrument recognition rather than specific defect detection.

The use of image-based methods, as opposed to real-time sensor-based systems, has been gaining traction due to its flexibility and the ability to integrate into existing workflows without intrusive modifications to the surgical process. Pontes for instance, demonstrated the use of adenosine triphosphate (ATP) sampling to monitor instrument cleanliness, but this approach lacks the ability to detect physical defects like corrosion or scratches. On the other hand, vision-based systems for instrument recognition, as shown by Liu et al. [27], achieved high accuracy in classifying instruments, although their focus was not on defect detection.

Further advancements in convolutional neural networks (CNNs), including models like YOLO (You Only Look Once), have shown promising results in real-time image-based defect detection and classification. For instance, Le et al. [28] successfully applied YOLOv8 for robust surgical tool detection, achieving a mean average precision (mAP50) greater than 95.6%. While this report is focused on the approach for instrument and defect classification, the potential for adapting YOLO models to defect classification is evident. Other studies have also highlighted the applicability of YOLO-based models for surgical instrument segmentation and detection [4], [17].

This report explores the various surgical instruments that were gathered as part of visual data gathering from industry partners. The instruments that were added were one of the best-sellers of our industry partners and contribute to a significant portion of the company's exports. Furthermore, this report outlines the development of a comprehensive dataset of instruments and defects based on visual data gathered from our industry partners, Daddy D Pro and Dr. Frigz. The dataset consists of a total of 2738 defected and 1676 undefected images and aims to provide a robust foundation for visual analysis and defect detection in the industry. The dataset covers a wide range of instruments commonly used in various applications, including medical, industrial, and research fields. By creating a well-structured and annotated dataset, we aim to facilitate the development of advanced computer vision algorithms for instrument recognition, defect identification, and quality control.

To ensure consistent quality and maintain its position in the global market, Pakistan's surgical instrument industry needs a technological solution [38]. This solution should focus on pre-shipment grading and inspection of manufactured instruments. The inspections must consider the regulations of various importing countries, such as ISO and FDA-GMP [6], [7]. These regulations require instruments to be free of defects like worn surfaces, cracks, and loose parts. Additionally, they need to be clean, free of stains and corrosion, and have movable parts that function smoothly. Some standards even recommend using magnification to detect tiny particles that might be missed by the naked eye [8]–[10].

The focus of this report is to highlight the importance of surgical instruments manufacturing industry for Pakistan as it has a major share in the country's exports. Moreover, identifying the defects in these surgical instruments is of utmost importance, as it can result in loss of business for manufacturer's and can put patient's life at risk as well [11]–[13]. Therefore, this survey represents a pivotal initiative in the realm of surgical instrument quality control, aiming to identify and address manufacturing defects with the expertise of our esteemed industry partners, Daddy D Pro and Dr. Frigz International. This report aims to identify various surgical instruments made in Pakistan and their associated manufacturing defects to ensure quality assurance. The document aims to provide a comprehensive overview of the diverse range of surgical instruments manufactured in Pakistan. Focusing on quality assurance, the analysis will delve into the identification of defects encountered during or after the manufacturing process. This exploration will serve as a valuable resource for manufacturers, healthcare professionals, and regulatory bodies, contributing to the improvement of surgical instrument standards [21]. Furthermore, the growing need for enhanced quality control in surgical instrument manufacturing and maintenance has led to the exploration of intelligent inspection systems. Manual inspection processes, traditionally reliant on human expertise, are inherently time-consuming, subjective, and prone to variability due to human fatigue or oversight. In medical contexts where precision and reliability are paramount, such as in surgical environments, even minor defects like corrosion, scratches, or micro-cuts can compromise the sterility and functional integrity of instruments, potentially resulting in serious clinical consequences including surgical site infections (SSI) [14]–[16], [20]. To mitigate these risks, the development of an Automated Optical Inspection (AOI) Tool offers a scalable, consistent, and efficient solution for real-time defect detection and classification.

## II. INDUSTRY PARTNERS

The pursuit of excellence in surgical instrument manufacturing has led to collaborative efforts between industry leaders and technological innovators. Two prominent companies, DaddyDPro and Dr. Frigz, have joined forces as industry partners in a groundbreaking project aimed at enhancing defect detection in surgical instruments in Pakistan. This collaborative initiative represents a paradigm shift, incorporating cutting-edge technology and expertise to elevate the quality standards of surgical instruments in the country.

### A. Company Profiles

1) Dr. Frigz:
   Dr. Frigz Figure 1 is a renowned manufacturer of surgical instruments based in Sialkot, Pakistan. With decades of experience, Dr. Frigz has earned a reputation for producing high-quality instruments for global healthcare markets. Dr. Frigz specializes in the traditional craftsmanship of surgical instruments while embracing modern manufacturing techniques to meet international standards.

2) Daddy D Pro:
   Daddy D Pro Figure 2 stands as a distinguished player in the surgical instrument industry in Pakistan, renowned for its commitment to quality and innovation. The company not only excels in traditional craftsmanship but also embraces cutting-edge technologies, including artificial intelligence (AI) and machine learning (ML), to enhance defect detection processes. Daddy D Pro is deeply involved in the production of surgical instruments, integrating technology to advance quality control measures [33].The collaboration between Daddy D Pro and Dr. Frigz revolves around implementing advanced technologies to enhance defect detection during the manufacturing process of surgical instruments in Pakistan. The project aims to address challenges related to quality control, reduce defects, and ensure that the instruments meet international standards.

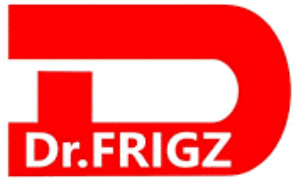

Fig. 1. Dr. Frigz International

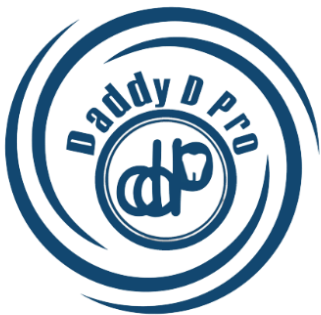

Fig. 2. Daddy D Pro

## III. An overview of Surgical Instruments and their Defects

Surgical instruments are tools that allow surgeons to open the soft tissue, remove the bone, dissect and isolate the lesion, and remove or obliterate the abnormal structures as a treatment [17], [36]. Surgical instruments are designed for use in various medical procedures, particularly during surgery. These instruments are crafted with precision to perform specific functions, and they play a critical role in ensuring the success of surgical interventions.

There are almost 14 surgical specialties recognized by American College of Surgeons, they range from orthopedic to vascular surgery [18], [34]. Each specialty has its own tools for specific procedures. New instruments are being launched every day to facilitate surgeries. Surgical tools are essential for medical procedures. Surgical instruments have a long history and they are believed to be originated a long time ago [19], [20], [35]. Since then, there has been continuous improvements in the design and usage of surgical devices and tools that are produced as a result of breakthroughs in science and technology. In the heart of surgical industry is Pakistan, particularly the city of Sialkot, known as a global hub for the production of high-quality surgical instruments.

The roots of surgical instrument manufacturing in Pakistan can be traced back several decades. Sialkot, with its skilled artisans and craftsmen, emerged as a center for producing precision instruments. Over the years, the industry has evolved, blending traditional craftsmanship with modern technologies to meet the increasing demands of the global healthcare sector. In the world of healthcare, the precision and reliability of surgical instruments is crucial for the success of medical procedures. The precision of surgical instruments during surgery of the utmost importance. However, with the passage of time, or at the time of manufacturing, these instruments are prone to defects that can hinder their performance [21]–[24]. These defects present considerable challenge for the manufacturers of such instruments to develop techniques and detection mechanism so that defects are detected before reaching the hands of surgeons and to prevent any compromise on patient's wellbeing.

### A. Classification of Instruments

The FDA in the United States categorizes medical devices into three classes—Class I, Class II, and Class III—based on their risk and significance. Examples of Class I devices encompass items like tongue depressors, bandages, gloves, bedpans, and basic surgical tools. Class II devices, which pose a moderate risk, include wheelchairs, X-ray machines, MRI machines, surgical needles, catheters, and diagnostic equipment [25]–[28]. Class III devices, reserved for higher-risk applications within the body, comprise heart valves, stents, implanted pacemakers, silicone implants, and hip and bone implants. Surgical instruments, crucial components in medical procedures, can be broadly categorized into disposables and non-disposables [29]–[32].

A majority of the world's surgical instruments are manufactured in specific cities and towns across Europe and Asia. Notable locations include Tuttlingen in Germany, Sialkot in Pakistan, Penang in Malaysia, Debrecen in Hungary, and Warsaw in Poland [38].

Surgical instruments are designed for performing specific tasks and they are used by nurses, surgeons and similar medical professionals. There are thousands of different types of surgical tools circulating in a hospital. According to [24], just one institution processed 2.6 million surgical tools, annually. According to [25], on an average, 6 trays of instruments are used for each surgery and each tray contains 38 surgical instruments on an average.

## B. Surgical Instruments in Pakistan

Manufacturers of surgical instruments in Pakistan predominantly produce Class I instruments and some Class IIb instruments [38].

The surgical instruments sector in Pakistan is characterized by a high degree of fragmentation and a pronounced focus on exports. This industry is primarily based in Sialkot, located in the Punjab province of Pakistan. It comprises numerous small and medium-sized manufacturers along with a handful of larger units. Operating primarily on an OEM (Original Equipment Manufacturer) model, manufacturers receive orders and specifications from international buyers, notably from Germany, the UK, and the USA.

Local manufacturers then make the products to meet the given specifications and export them to overseas buyers, who often brand the instruments and distribute them through international distributors. Over the past decade, Pakistan's market share in surgical instruments has demonstrated relative stability at 0.7, indicating a robust correlation of 0.96 with global demand [38].

The surgical instruments industry is predominantly concentrated in and around the outskirts of Sialkot, with over 99% of the country's production emanating from this region. This sector comprises a total of more than 2300 companies, with approximately 30 being categorized as large enterprises, while the rest are distributed among 150 medium-sized units and numerous small-scale units. The industry demonstrates robust productivity, churning out an average of over 150 million pieces annually, valued at around Rs 22 billion. Notably, more than 95% of this production is earmarked for exportation. Classified under the light engineering industry, this sector has carved a niche for itself with specialized skills and a stable market share in exports.

In addition to the small and medium-sized units, a handful of larger establishments operate with a 90% integrated system. While the majority of these larger and medium-sized firms engage in exporting, the smaller or vendor units typically cater to commercial exporters or traders. The primary raw material utilized in production is 'steel,' with approximately 60% of it sourced locally, and the remaining 40% imported, primarily from Germany.

From a trade perspective, Pakistan's exports fall into four broad categories. These categories encompass (i) HS Code 9018 – Instruments for medical, surgical, and dental; (ii) HS Code 9021 – Orthopedic appliances; (iii) HS Code 9022 – Equipment utilizing X-rays, alpha, beta, gamma rays. Notably, the majority of Pakistan's exports fall under the category 9018, emphasizing the country's significant presence in the international market for medical instruments [39].

## C. HS Codes for Surgical instruments

HS (Harmonized System) is an international standard for products classification. The HS code for surgical instruments is HS 9018. An overview of the surgical instruments falling in this category is shown in Table I

TABLE I
HS CODES FOR SURGICAL INSTRUMENTS

| HS Codes | Instruments |
|----------|-------------|
| 901811 | Electro-Cardiographs |
| 901812 | Ultrasonic scanning instruments |
| 901813 | Magnetic imaging instruments |
| 901814 | Scintigraphic instruments |
| 901819 | Electric-diagnostic instruments |
| 901820 | Infra-red ray instruments |
| 901831 | Syringes and needles |
| 901832 | Suture needles and Tabular metal |
| 901839 | Cannulas, Catheter, Needles, etc. |
| 901841 | Drill Engines for Dentistry |
| 901849 | Dental science apparatus |
| 901850 | Ophthalmic appliances and apparatus |
| 901890 | Medical, veterinary, etc. instruments |

## D. Pakistan's Surgical instrument trade

Surgical instrument's trade makeup a sizable portion of country's exports. The statistics of the surgical instrument's exports of Pakistan from 2015 to 2022 [38] and [40] can be seen from Table II

TABLE II
PAKISTAN SURGICAL TRADE FROM 2015–2022

| Year | Exports in USD (Million) | % Share of Pakistan Globally |
|------|--------------------------|------------------------------|
| 2022 | 307.7 | 0.6% |
| 2021 | 426.0 | 0.7% |
| 2020 | 361.3 | 0.6% |
| 2019 | 405.5 | 0.7% |
| 2018 | 375.5 | 0.7% |
| 2017 | 361.1 | 0.7% |
| 2016 | 326.0 | 0.7% |
| 2015 | 332.6 | 0.7% |

Pakistan boasts a diverse range of surgical instruments catering to various medical specialties. Scalpels, forceps, scissors, needle holders, retractors, and dental instruments are among the myriad tools meticulously crafted by skilled artisans. These instruments serve as extensions of the surgeon's hands, designed for specific functions to ensure precision and efficiency in the operating room. The craftsmanship involved in the production of surgical instruments in Pakistan is a testament to the dedication of local manufacturers. The manufacturing processes are characterized by attention to detail, the use of high-quality materials, and adherence to international quality standards. Artisans employ traditional techniques alongside modern innovations, ensuring the durability and precision of each instrument.

The impact of Pakistani surgical instruments extends far beyond national borders. Sialkot has become a key player in the global market, exporting instruments to medical facilities worldwide. The instruments manufactured in Pakistan are renowned for their quality, meeting the stringent standards set by international regulatory bodies. This has not only contributed to the country's economic growth but has also positioned Pakistan as a reliable and competitive player in the global healthcare supply chain.

### E. Types of Surgical instruments

Surgical instruments are categorized into various groups based on their functions, applications, and usage in medical procedures. These categories provide a systematic classification to help healthcare professionals, manufacturers, and regulatory bodies understand and organize the diverse array of instruments used in surgery. Table III includes a brief overview of the surgical instruments categories and their associated instruments.

TABLE III
SURGICAL INSTRUMENTS AND THEIR CATEGORIES

| Category | Instruments |
| --- | --- |
| Cutting instruments | Scalpel, scissors, Biopsy punches, etc. |
| Grasping and Holding instruments | Forceps, Towel Clamps, etc. |
| Clamping and Occluding Instruments | Homeostat Forceps, Vascular Clamp, etc. |
| Retractors | Self-retaining, hand-held retractors, etc. |
| Probing and Dilating instruments | Dilators, Urethral probes, etc. |
| Suture and Stapling instruments | Staplers, Needle holders, etc. |

Since, the industry partners of this project have specialization in manufacturing surgical instruments especially dental products, the consequent sections of the survey will focus on surgical and dental instruments in the light of data provided by our industry partners Dr. Frigz International and Daddy D Pro. The categorization of dental instruments can be seen in Table IV

TABLE IV
DENTAL INSTRUMENTS AND THEIR CATEGORIES

| Category | Instruments |
| --- | --- |
| Examination Instruments | Dental mirrors, probes, and explorers, etc. |
| Diagnostic Instruments | Periodontal probes, dental radiography instruments, etc. |
| Surgical Instruments | Dental forceps, elevators, and surgical curettes, etc. |
| Periodontal Instruments | Scalers, curettes, and periodontal knives, etc. |

### F. Surgical Instruments manufactured in Pakistan

Some of the famous instruments that are largely exported from Pakistan, also largely manufactured by our industry partners are explained in detail below:

*1) Scalpel:* It is used for cutting tissue and initial incision. It consists of a handle and blade. It is often referred by its blade number that indicates the degree of sharpness or its purpose [37]. Figure 3 shows the various types of scalpel blades and their uses. The different types of blades are used for different operations. 10- Blade is used for making large skin incisions like in laparotomy. 11- Blade is used for sharply angled incisions. 15- Blade is used for finer incisions as it is smaller version of 10- Blade.

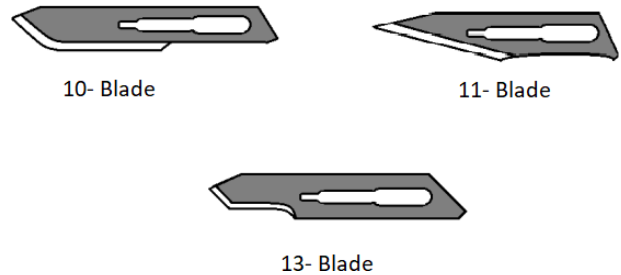

Fig. 3. Scalpel Blades

*2) Scissors :* They are used for suture, cutting tissue, and for dissection. They can be curved and straight and can be used for cutting finer or heavy structures. Figure 4 shows different kinds of scissors. The Mayo scissors are available in many varieties. The curved are used for cutting heavy tissue and straight are used for cutting suture. Metzenbaum scissors are used for cutting thin tissue like the heart and also for blunt dissection. Pott's scissors are famously used for creating incisions in thin vessels like the blood vessels. Moreover, Iris scissors are used for cutting fine suture and fine dissection.

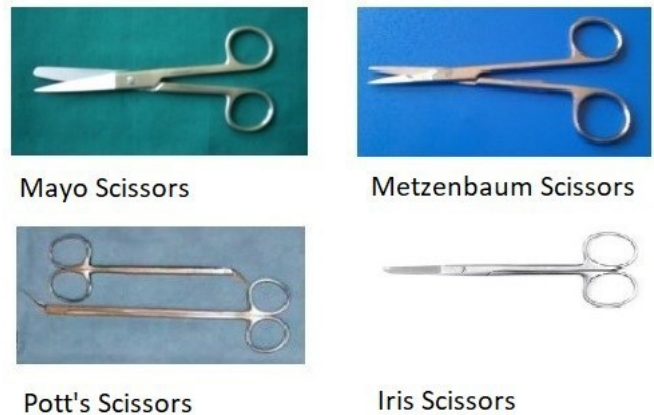

Fig. 4. Surgical Scissors

*3) Forceps:* They are also known as thumb forceps, pickups, grasping or nonlocking forceps. They are used for grasping objects and tissue [37]. They can have serrated and nonserrated tips. Figure 5 shows different types of forceps. Tissue forceps are non-toothed and are commonly used for traction and fine handling tissue during dissection. Adson Forceps on the other hand, have toothed tips and are used in handling dense tissue like the skin closures.

Bonney forceps are used for holding heavy tissue like the fiscal closure. DeBakey Forceps are mostly used for grasping atraumatic tissue during dissection. Also, Russian forceps are used for the same purpose as DeBakey's as can be seen from Figure 6.

Another type of Forceps used in Dentistry are called Extraction Forceps that are primarily for extracting tooth. There are different types of Forceps for different uses and they are

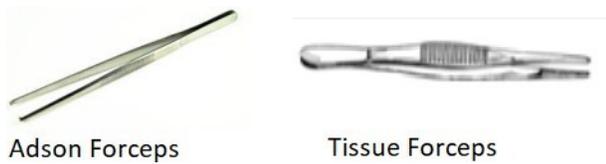

Adson Forceps    Tissue Forceps

Fig. 5. Tissue and Adson Forceps

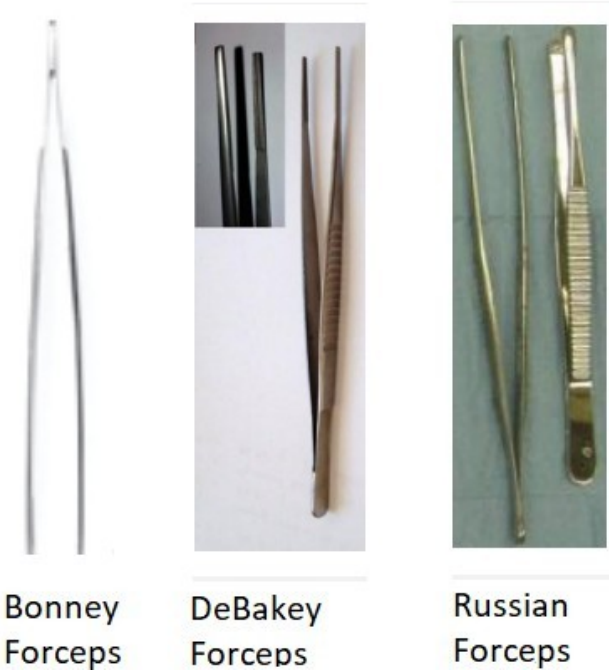

Bonney
Forceps    DeBakey
Forceps    Russian
Forceps

Fig. 6. Bonney, DeBakey and Russian Forceps

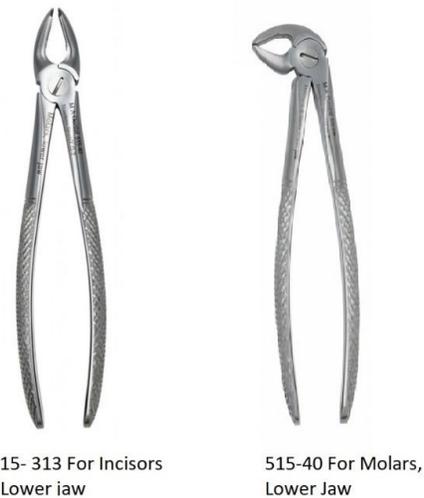

515- 313 For Incisors
- Lower Jaw    515-40 For Molars,
Lower Jaw

Fig. 7. Extraction Forceps

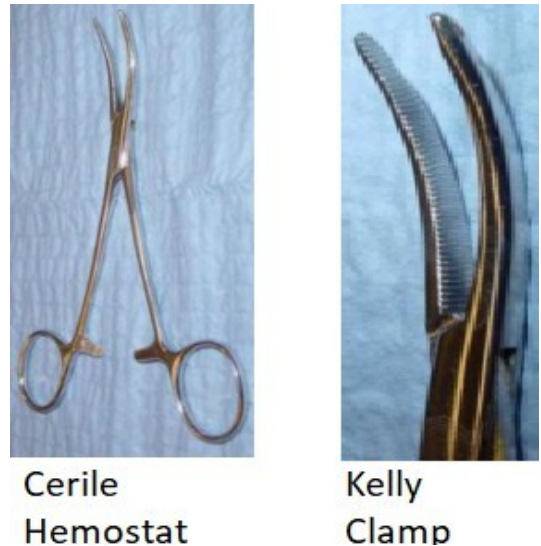

Cerile
Hemostat    Kelly
Clamp

Fig. 8. Cerile Hemostat and Kelly Clamps

mostly numbered to identify their usage. They can be seen in Figure 7 below.

*4) Clamps:* They are also called locking forceps and are used to hold objects or tissue during surgery and provide hemostasis. They can be atraumatic or traumatic. Figure 8 shows different types of clamps. Crile Hemostat also called the non-toothed clamp is used in blunt dissection and to hold tissue to be tied off. Kelly clamp is a famous surgical clamp that is used for grasping larger blood vessels or tissues.

Kocher clamp on the other hand, is used to hold tissue to be removed. Allis and Babcock clamps are used for holding intestines during dissection and have slightly rounded jaws as can be seen from Figure 9 below.

*5) Needles and Sutures:* Needles are available in different cutting edges and shapes for various applications. Whereas, Suture are available in different sizes and can be absorbable/non-absorbable. Figure 10 shows different types of needles. Needles come in various shapes and sizes and are selected based on the application. The primary job of needle is to pass suture for that it must dissect through tissue. The two famous needle types are "Conventional cutting needle" and "Tapered needle". Conventional cutting needles are usually triangular with really sharp edges and are used for tougher tissues like the skin. Whereas, the Tapered needles are used mostly used with softer tissues like the intestine.

Figure 11 shows a surgical Needle holder.

The shape of needle also plays a crucial role in the specialized procedures. They are selected based on the necessity and complexity of the surgical procedure. They are commonly found in the form of semi-circles, straight ones, J shape, and compound curved. Curved ones are generally used in surgical procedures whereas straight ones are used suturing skin. Sutures are surgical needles and are generally categorized into two main types. The first types are non-braided and braided. The second types are non-absorbable and absorbable. Moreover, Sutures can be made of synthetic or natural materials. Figure 12 shows different Suture types.

*6) Retractors:* Retractors are needed to hold back tissue, other objects, or an incision open during surgical procedures.

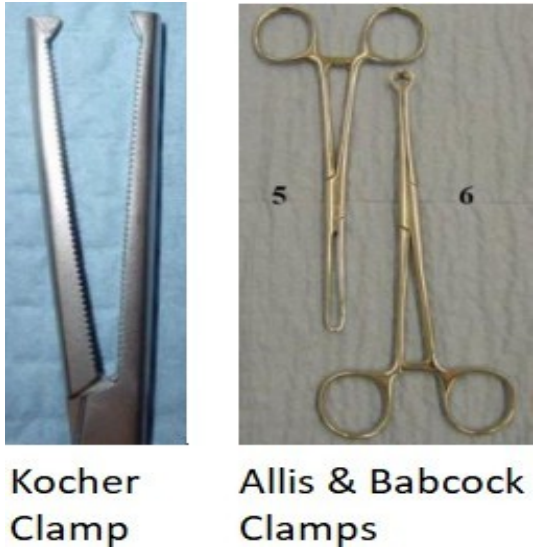

Fig. 9. Kocher and Allis and Babcock Clamps

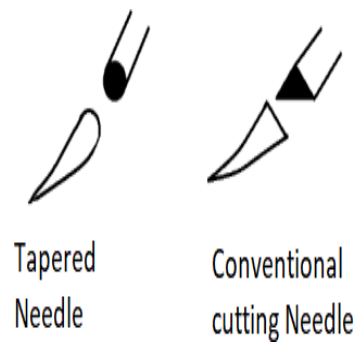

Fig. 10. Different Needle types

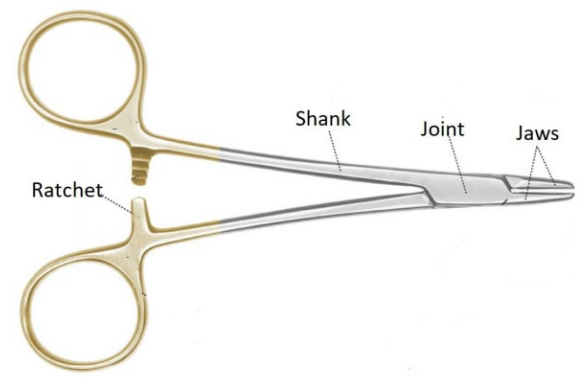

Fig. 11. Needle holders

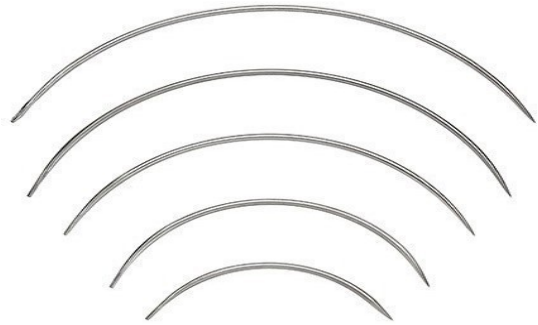

Fig. 12. Suture Needles

They are either self-retaining or hand-held. Figure 13 shows different types of retractors. Deaver Retractor is mostly used to hold thicker tissue like the abdominal wall. Army-Navy retractor is commonly used to gain exposure to layers of skin. Whereas Richardson retractor is majorly used in holding back complex deep tissue structures.

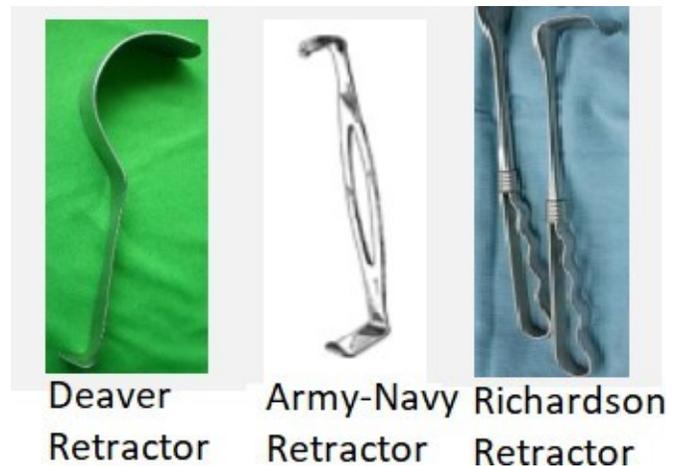

Fig. 13. Different types of Retractors

Weitainer retractor also called as "Wheaty" is used to gain exposure to small surgical sites. On the other hand, the Bookwalter retractor is a self-retaining system commonly anchored to surgical operating table as seen in Figure 14.

*7) Curettes:* A curette it is a stylus like surgical instruments used to remove or clean any excess of blood or infected tissues using a scraping or scooping motion. Dental curettes are used to remove tartar deposits form teeth. A curette may be double or single-ended, and it may have spoon like ends, hooks, or gouges. Gracey curettes as shown in Figure 15 are designed with a single cutting edge, these curettes are specific to either the mesial or distal surfaces of teeth. They are Subdivided into various numbers (e.g., Gracey 1/2, 11/12) to accommodate different tooth surfaces and locations as can be seen from Figure 15.

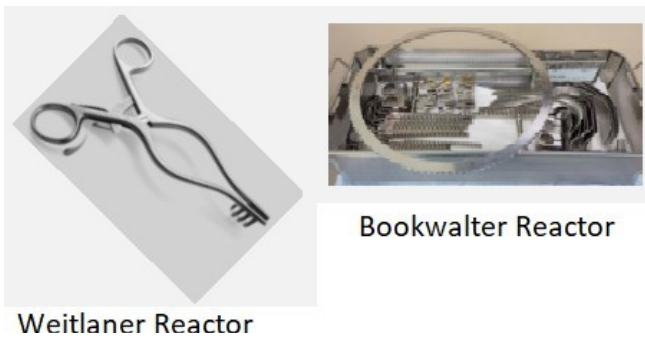

**Bookwalter Reactor**

**Weitlaner Reactor**

Fig. 14. Weitlaner and Bookwalter Reatractors

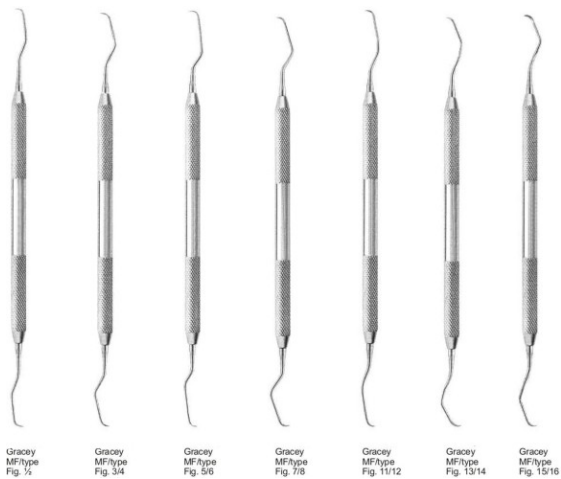

Fig. 15. Gracey Curettes

## G. Defects

Surgical instruments must undergo a rigorous process of cleaning, sterilizing and inspection after surgeries to ensure their proper maintenance, increase longevity of tools and prevent health hazards due to contaminated tools. A critical step involved in doing so is the visual inspection conducted by medical staff. The accounting of tool kits is often a labor intensive and mundane task which could be easily automated. Feedback from stakeholders indicate that human errors in monitoring and inspection do occur occasionally, resulting in patient harm if unidentified or loss of time during surgery if identified by surgeons. This can be attributed to subtle differences and small features of tools as well as the strenuous effect on the eyes. Defects in surgical instruments pose significant dangers due to their potential to compromise critical aspects of medical procedures and patient safety. The inherent risks associated with these defects can have severe consequences, and understanding their dangers is essential for maintaining the highest standards of healthcare. The key reasons why defects in surgical instruments are dangerous include:

- Impact on Precision and Accuracy: Surgical procedures require an exceptional level of precision. Defects such as sharp edges, deformities, or irregularities can impede the surgeon's ability to execute precise movements, increasing the risk of unintended tissue damage or complications.
- Infection Risks: Instruments with defects, especially those involving pores or corrosion, can harbor bacteria and other pathogens. This significantly elevates the risk of post-surgical infections, which can lead to extended recovery periods, additional medical interventions, or even life-threatening complications.
- Compromised Structural Integrity: Defects weaken the structural integrity of surgical instruments. This compromises their durability and increases the likelihood of breakage during procedures. Broken instruments may leave fragments in the patient's body, necessitating additional surgeries for retrieval.
- Functional Impairment: Functional deficiencies in surgical instruments can hinder their intended use. When instruments fail to perform as expected, surgeons may face challenges in achieving optimal outcomes during procedures, potentially resulting in incomplete surgeries or the need for additional interventions.
- Patient Safety at Risk: Surgical instruments are vital tools in ensuring patient safety during medical procedures. Any compromise in the quality of these instruments, whether due to defects in design, manufacturing, or wear and tear, directly jeopardizes the safety of patients undergoing surgery.
- Long-Term Consequences: Defects in implanted devices, such as stents or pacemakers, can have long-term consequences. Malfunctioning implants may require corrective surgeries, leading to increased healthcare costs, prolonged recovery periods, and potential psychological impacts on patients.
- Legal and Ethical Implications: The discovery of defects in surgical instruments can result in legal and ethical challenges for healthcare providers and manufacturers. Medical malpractice claims, damage to reputations, and regulatory consequences may arise, creating a complex and contentious environment.
- Financial Implications: Surgical complications resulting from defective instruments can lead to increased healthcare costs. Patients may require extended hospital stays, additional medical treatments, and rehabilitation, contributing to a financial burden on both individuals and healthcare systems.

Surgical defects refer to imperfections, irregularities, or shortcomings in surgical instruments that can compromise their functionality, durability, and safety during medical procedures. These defects can arise during the manufacturing process or develop over time due to various factors. Common surgical defects include pore formation, corrosion, cuts, and other issues that may affect the instrument's performance.

This part of the survey aims to comprehensively explore and analyze the prevalence, types, and consequences of defects in surgical instruments. By delving into this critical aspect of healthcare, we endeavor to shed light on the multifaceted dimensions of instrument defects, ranging from manufacturing flaws to wear-and-tear issues arising from prolonged usage. Understanding the nuances of these defects is essential for fostering improvements in design, manufacturing processes, and maintenance protocols.

*1) Types of Defects:* Different types of defects associated with surgical instruments are described in detail below.

- Pores:

A broken pore indicates that there is rust and the instrument will not perform adequately. Figure 16 shows broken pore on the surface of Adison Forceps.

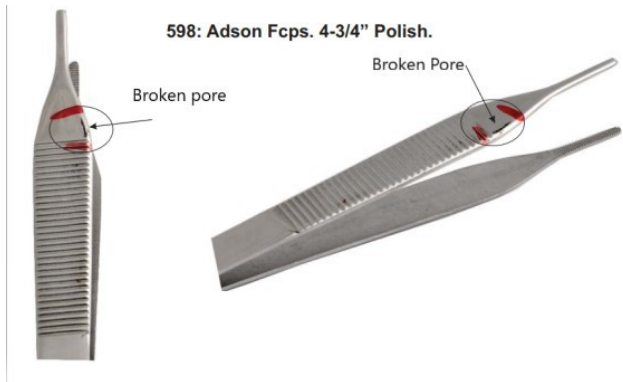

Fig. 16. Pores on Adson Forceps

Figure 17 shows broken pore on the probe of Lester bandage scissor. The instruments should be carefully inspected so that the surface of instruments is shiny, and unremarkable.

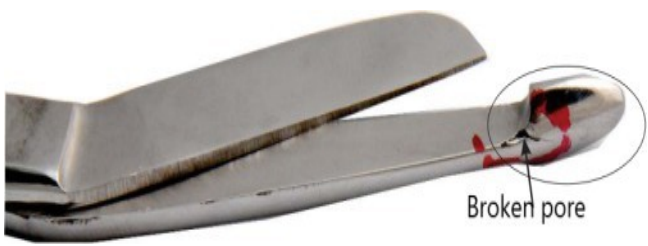

Fig. 17. Broken pore on Lester Bandage scissors

Figure 18 shows the broken pore on the inner side of the Forceps. During inspection of the Forceps, it should be kept in mind that the inner handle of the forceps doesn't have any remarkable defect.

- Crack:

While inspecting Nail cutters it should be kept in mind that its surface is shiny and unremarkable. The instrument should be oiled properly and it should open and close

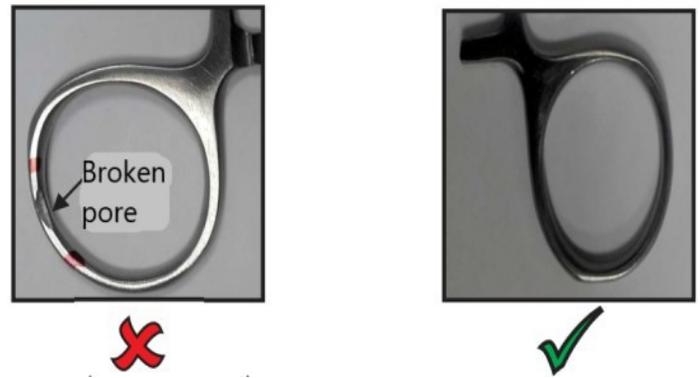

Fig. 18. Broken pore on Forceps

with ease. Figure 19 depicts the crack on the tip of Nail cutter.

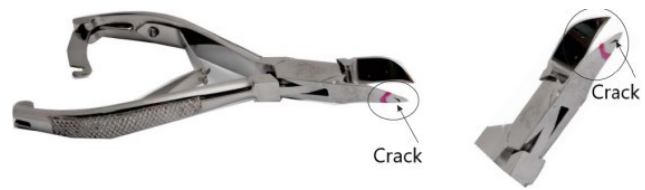

Fig. 19. Crack on Nail cutter

Cracks are the most common defects to occur in surgical instrument an example is that of a scissor as shown in Figure 20.

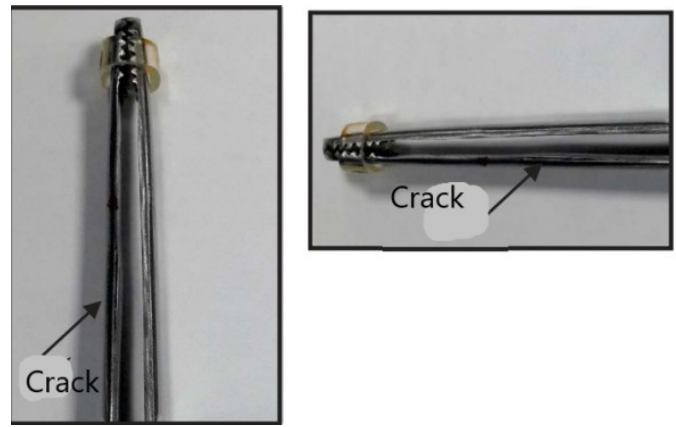

Fig. 20. Crack on Scissors

Another example of crack occuring on the surface of Needle holder is shown in figure 21.

- Corrosion/Rust:

Rust/Corrosion accumulates on surgical instruments due to a number of factors. It could either be poor packaging from the manufacturer's end or a result of prolonged use and exposure to rust causing elements/factors. Rusted instruments should be taken out during inspection as they can hinder in performance of instruments. Figure 22 shows rusted Needle holders. Whereas Figure 23 shows the rust on tip of needle holders and other parts.

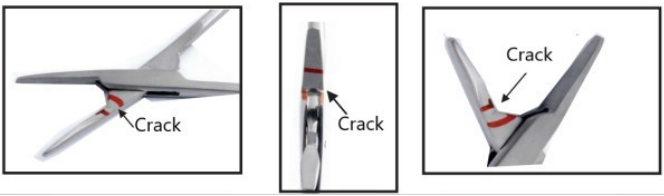

Fig. 21. Crack on Needle holders

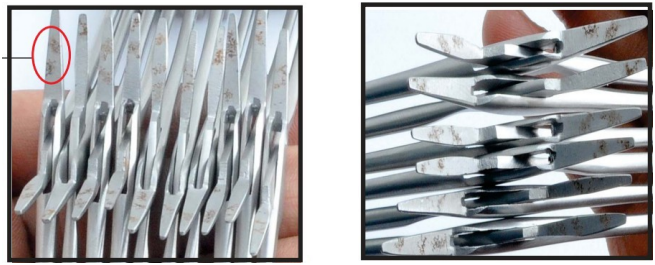

Fig. 22. Corrosion on Needle holders

Preventing surgical defects involves stringent quality control measures during the manufacturing process. This includes using high-quality materials, employing precise manufacturing techniques, implementing rigorous inspection processes, and adhering to industry standards. Regular maintenance, proper sterilization methods, and careful handling by healthcare professionals also play crucial roles in preventing defects and ensuring the longevity of surgical instruments. Continuous improvement through feedback mechanisms and technological advancements further contributes to defect prevention in the surgical instrument industry.

## IV. Visual Data Gathering and Curation

The success of any computer vision-based defect detection system is deeply rooted in the quality and relevance of its underlying dataset. In the domain of medical imaging, particularly for surgical instrument inspection, the diversity, resolution, and defect annotation of visual data are critical for training robust and generalizable deep learning models. Visual data gathering serves as the foundation for supervised learning approaches, enabling models to differentiate between defective and non-defective instrument surfaces under varied clinical conditions.

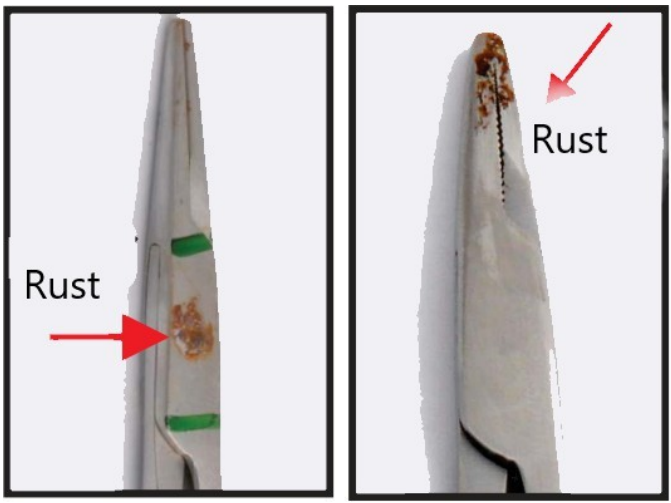

Fig. 23. Rust on Needle holders

Surgical instruments present a unique challenge due to their reflective metallic surfaces, complex geometries, and the subtlety of many defect types such as corrosion, microscratches, or pores. To build a dataset capable of representing these nuances, collaboration with industry partners is essential. In this project, visual data was collected in partnership with surgical instrument manufacturers—including Daddy D Pro—who provided access to a wide range of instruments under real-world use conditions. This industrial collaboration ensured the dataset reflects operational variability and includes domain-relevant defect cases.

Controlled imaging conditions were established using standardized setups involving consistent lighting, camera calibration, and fixed orientations, aligning with established best practices in medical AI imaging pipelines. High-resolution images were captured across multiple instrument types, with expert-led manual annotation of defect classes to ensure clinical validity. Such collaborative and standardized approaches have proven effective in recent medical AI studies, especially in fields like endoscopy and pathology, where image fidelity and annotation precision are pivotal.

By gathering high-quality visual data directly from manufacturing sources, this project addresses the data scarcity and heterogeneity issues prevalent in surgical AI research, paving the way for reproducible and clinically relevant defect classification systems.

### A. Surgical Instruments from Industry Partner

Focusing on the analysis of surgical instrument data obtained from two key industry partners: Dr. Frigz International and Daddy D Pro. The data encompasses a comprehensive collection of information pertaining to both defective and non-defective instruments produced by these companies. This data analysis will provide a critical foundation for collaborative efforts with our industry partners. We can utilize the insights to propose improvements that lead to:

- Reduced instrument defects: Lowering the incidence of defects ensures a higher quality product reaching hospitals and surgical teams.
- Enhanced production efficiency: Identifying and addressing root causes of defects can streamline manufacturing processes.
- Improved patient safety: Flawless instruments are crucial for safe and successful surgical procedures.

Surgical instruments such as Surgical probes, Scalpel, Scissors, and Curettes make up a sizeable portion of the overall surgical instruments manufactured by our industry partners Dr. Frigz International and Daddy D Pro. Below are the details of these instruments, their associated categories and how dataset was formulated from them. The following sections delve deeper into the collected data, exploring the details of instruments from both Dr. Frigz International and Daddy D

Pro. We will analyze trends and patterns in defect occurrences, ultimately formulating actionable recommendations to elevate the quality and reliability of surgical instruments produced by our industry partners.

## V. Data Collection and Pre-processing

Data curation is a critical phase in the development of a robust and accurate machine learning model. In the context of this project, the goal of the data curation process was to gather high-quality images of surgical instruments that accurately represent the defects we aim to classify, such as corrosion, scratches, cuts, and pores. Given the complexity and importance of defect classification in surgical tools, the data curation process was designed to ensure consistency, diversity, and relevance across the dataset.

### A. Instrument Setup

A controlled imaging setup was implemented to minimize environmental variability and ensure consistent, high-quality data for defect detection. The setup included a Canon EOS 250D camera (1600x1600 resolution), a tripod stand for fixed positioning, a Puluz photo light box for uniform lighting, and a green background to enhance contrast with surgical instruments and reduce visual distractions. This configuration ensured that the YOLOv8 model could focus on identifying defects without interference from lighting or background inconsistencies.

### B. Surgical Instruments in Dataset

The initial dataset included five surgical instruments (Carver, Ex-Probe, Probe, Scalpel, Scissors) as shown in Figure 24, each with defective and non-defective samples captured under a controlled setup.

Defects such as corrosion, cuts, scratches, and pores were documented to create a balanced dataset for model training. To improve dataset diversity and enhance model generalization, six additional instruments were later added—Bandage Scissors, Dressing Forceps, Mcindoe Forceps, Nail Clippers, Teale Vulsellum Forceps, and Uterine Curettes—bringing the total to 11 instruments as shown in Figure 25.

### C. Image Annotation

Each image was carefully reviewed by domain experts that were provided by our industry partner Daddy D Pro, who labeled the type of defect (e.g., corrosion, scratches, cuts, pores) or marked it as non-defective. This manual annotation process was meticulous to ensure that the labels were accurate and that no defective area was missed.

This curated dataset serves as the backbone of the defect classification system, ensuring that the model has a sufficient amount of labeled data to learn from. It also allows the model to generalize well, as it includes a wide range of instruments and defect types, which reflect real-world conditions that the system will encounter once deployed in a clinical environment.

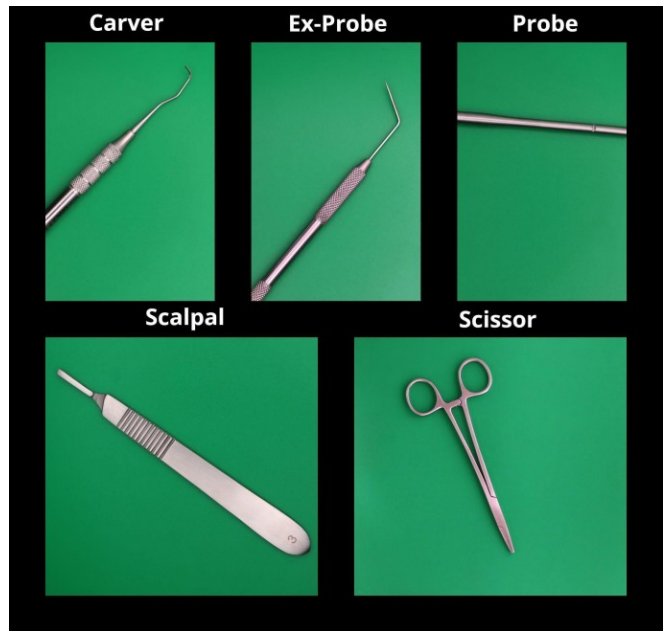

Fig. 24. Initial Surgical Instruments in Dataset

### D. Data Augmentation

Data augmentation plays a vital role in enhancing the generalizability and robustness of machine learning models, especially in tasks where the dataset is limited or prone to overfitting. In the context of this project, augmenting the dataset of surgical instruments was essential to ensure that the model could accurately classify defects across a variety of real-world scenarios.

he main motivation behind applying data augmentation in this project was to simulate real-world inspection conditions under which surgical instruments are examined for defects. These conditions include variations in the orientation of instruments, differences in lighting environments, and the presence of minor wear or noise due to usage or image quality. Instruments may be viewed from multiple angles during inspection, and the model needs to detect defects regardless of their positioning. Similarly, defects like corrosion or scratches can appear differently under varying lighting, necessitating robustness in the model's detection capabilities. Additionally, real-world images may contain slight imperfections, such as surface wear or noise, which the model should handle effectively without performance degradation.

To address these challenges, several augmentation techniques were employed, each designed to enhance the model's ability to generalize across diverse scenarios while preserving the essential characteristics of the defects. Rotation was applied at fixed angles (90°, 180°, 270°) and at random within a ±20° range to ensure orientation invariance, which is critical for instruments placed differently during inspection. This technique is supported by findings from Shorten and Khoshgoftaar [21], who demonstrated that rotation improves model generalization, particularly in datasets where object orientation varies. Brightness and contrast adjustments of ±20% were used to replicate different lighting conditions, enabling the model to recognize defects regardless of external illumination. Gaussian noise was added to simulate real-world imperfections from low-quality cameras or aged instruments, thereby improving the model's robustness. Lastly, unsharp masking was applied to enhance image edges, making subtle defects more prominent and improving the model's focus on relevant defect features. Theses augmentations can be seen from Figure 26. Collectively, these augmentation strategies ensured the model's resilience and reliability in practical defect detection tasks.

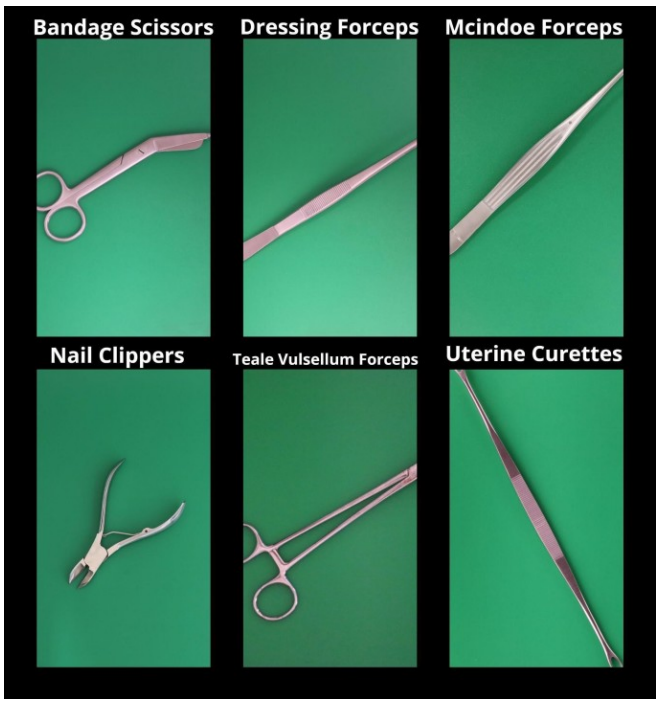

Fig. 25. Additional Surgical Instruments in Dataset

## VI. Data Packaging

To ensure efficient utilization of the collected surgical instrument dataset for defect classification, systematic data packaging was critical. Proper data formatting, splitting, and storage enable streamlined training and evaluation of deep learning models, particularly YOLOv8, which requires specific directory structures and annotation formats. This section outlines the steps taken to prepare the dataset, including annotation conversion, YAML configuration, data splitting, and storage/access protocols, all tailored to meet the requirements of the YOLOv8 model.

### A. Data Splitting

To evaluate model performance effectively and prevent overfitting, the dataset was divided into training, validation sets:

- Training: 80%
- Validation: 20%

A stratified random split was employed to ensure balanced distribution of defective and non-defective samples across all sets. This method maintains class proportion, ensuring the model is exposed to all classes during training and evaluation.

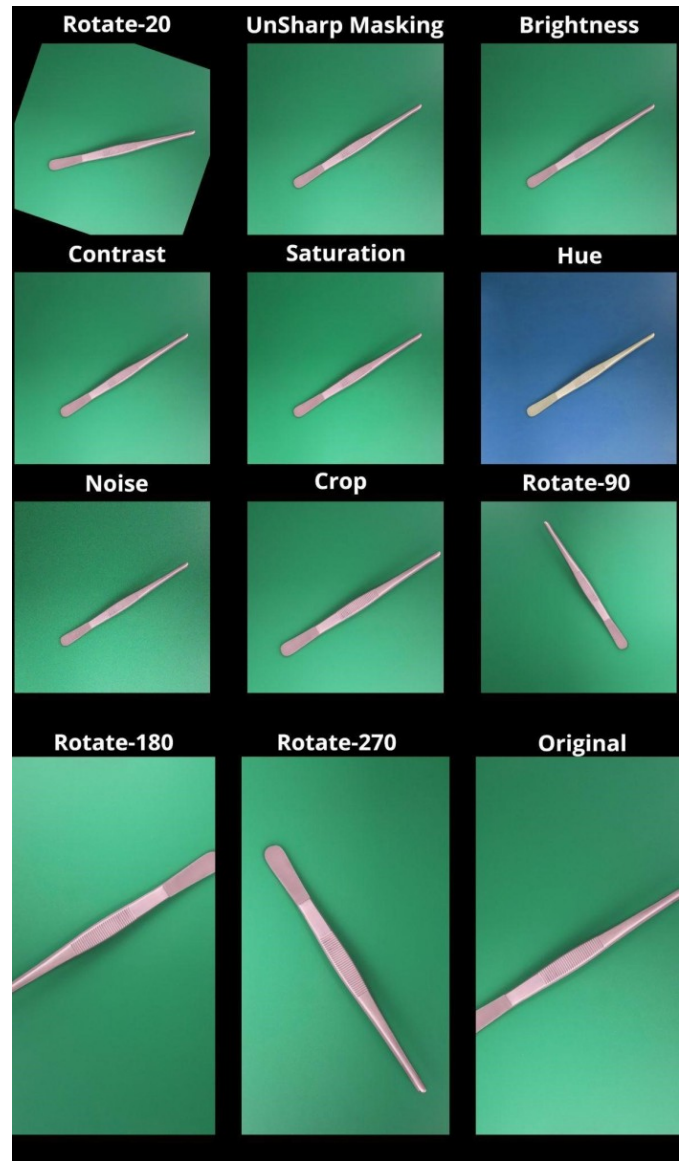

Fig. 26. Overview of Augmentations on Surgical Instruments

The split was performed using Python (e.g., sklearn.model_selection) with custom logic to handle augmentation data. Only original and augmented images were included in the training set, while validation and test sets consisted solely of original, unaugmented samples.

### B. Format Conversion for YOLOv8

YOLOv8 mandates a specific annotation format for object detection tasks, where each image must have an accompanying `.txt` file containing normalized bounding box coordinates in the form:
`<class-id> <x-center> <y-center> <width> <height>`.
All values are normalized with respect to the image width and height.

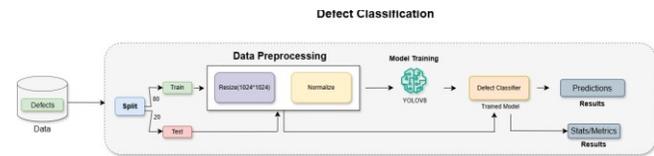

Fig. 29. Overview of Defect Classification

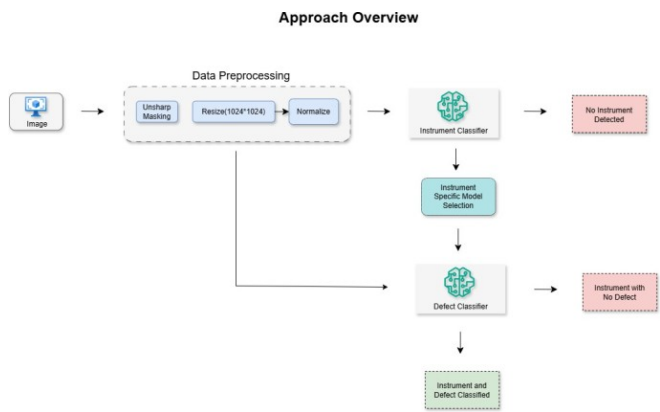

Fig. 27. Overview of Approach

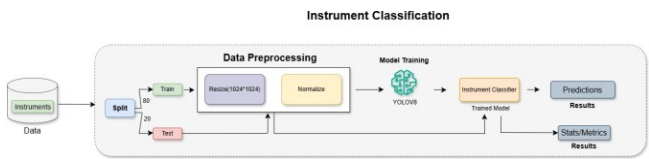

Fig. 28. Overview of Instrument Classification

These annotations were converted using custom Python scripts leveraging the "Element Tree" module. The conversion process involved extracting the bounding box coordinates (`xmin`, `ymin`, `xmax`, `ymax`), computing the center points and dimensions, and subsequently normalizing them. Each annotation was saved in a `.txt` file named identically to its corresponding image, ensuring compatibility with YOLOv8's expected directory structure. All label files were organized under the `labels` folder, mirroring the hierarchy of the `images` directory. Dataset is hosted on a private GitHub Repository.

## VII. Development of AI/ML based Approach

In the approach for defect classification in surgical instruments, we employed a two-stage classification process: first identifying the type of instrument, followed by a specific model trained to detect defects on that instrument. This process was crucial for ensuring accuracy and efficiency in defect detection, as each instrument may have different characteristics that influence both the appearance of defects and the type of defects it is prone to as can be seen in Figure 27 below.

### A. Instrument Classification

The first step in defect detection is to classify the type of instrument from the image. This classification is necessary because different surgical instruments exhibit varying structures, textures, and forms, which can affect how defects manifest as can be seen from Figure 28. Instruments like forceps, scissors, and scalpels have distinct shapes and surfaces, and the nature of defects like corrosion, scratches, or cuts can vary significantly between instruments. For example, defects like corrosion may appear prominently on metallic surfaces of instruments like scissors, while a porous surface on certain forceps may resemble a defect to a general-purpose model. Therefore, accurate instrument classification helps guide the system toward an instrument-specific defect model that is tailored to the defects typically associated with that instrument. We used YOLOv8 for instrument classification due to its high performance in real-time object detection and classification. YOLOv8's speed and accuracy made it ideal for surgical environments, where timely decisions are crucial. By first identifying the instrument type, we ensured that the system could direct the next stage of classification to the appropriate defect model for that specific instrument, minimizing false positives or misclassifications.

### B. Instrument-Specific Defect Classification

Once the instrument is classified, we move to the second stage: defect detection using a model specifically trained for that instrument. Each instrument type has unique structural properties and is prone to different types of defects. Therefore, instead of using a single defect detection model for all instruments, we developed instrument-specific models to improve detection accuracy as seen in Figure 29. For instance, an instrument like a scissor has sharp, metallic blades where defects like corrosion or cuts are likely to occur. In contrast, an instrument like a forcep may have textured surfaces or multiple joints, where defects like pores or scratches can manifest. Applying the same defect detection model to all instruments could lead to errors due to these structural differences.

By using instrument-specific models, we were able to:

- Tailor the detection process to each instrument's unique structure, ensuring that features like surface texture or the presence of joints did not interfere with the detection of true defects.

- Increase the accuracy of detecting defects that are more common in certain instruments. For example, corrosion is more likely to occur in metallic instruments, while pores may appear in instruments with more complex surfaces like forceps.

- Avoid confusion that might arise from the instrument's own structural features being misclassified as defects. Additionally, some instruments exhibit symmetry or specific surface characteristics that could be misinterpreted by a general-purpose model. For instance, certain forceps have naturally occurring pores or patterns that might be identified as defects by a non-specific model. To avoid this, we used specific models that learned to differentiate between the intended features of the instrument and true defects.

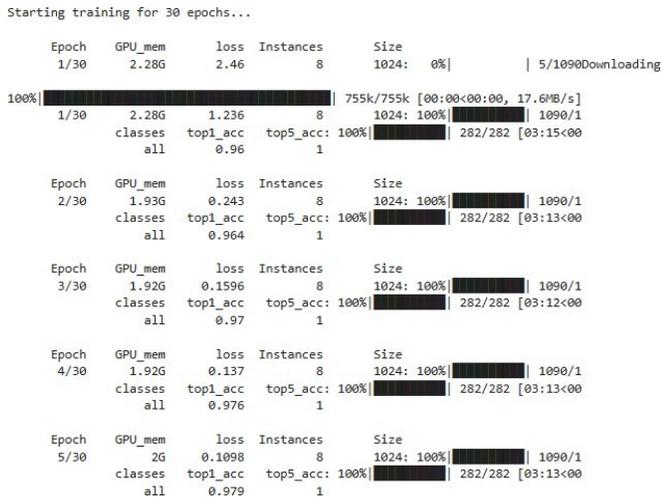

Fig. 30. Overview of Model Training

## VIII. Model Training and Performance

Each instrument-specific defect detection model was trained on a dataset containing both defective and non-defective examples of that instrument. This ensured that the model learned to distinguish between different types of defects (such as corrosion, cuts, scratches, and pores) and could accurately classify defects when presented with new images. The instrument-specific approach yielded high accuracy across all instruments, as shown in the averaged accuracy results for each defect type in instruments like Bandage Scissors, Dressing Forceps, Mcindoe Forceps, and Uterine Curettes in Figure 30. The model's ability to focus on the unique defect features of each instrument led to greater precision and fewer misclassifications compared to a single generalized defect detection model.

The two-stage process of instrument classification followed by instrument-specific defect detection allowed us to achieve high accuracy in identifying defects in surgical instruments. By customizing the defect models to each instrument, we ensured that the system could detect even subtle defects, while minimizing false positives caused by the instrument's own structural features. This approach is not only scalable but also provides a high degree of reliability in maintaining instrument integrity and ensuring patient safety in surgical settings.

## IX. Comparative Analysis

In this project, we trained several state-of-the-art models—EfficientNet-b4, Resnet-152, Resnext-101, YOLOv8, and YOLOv11—on the same dataset using identical hyperparameters to conduct a comparative analysis. The goal of this analysis was to determine the most effective model for the task of defect classification in surgical instruments. This systematic comparison allowed us to evaluate the strengths and weaknesses of each model in terms of accuracy, speed, and overall performance under identical conditions, ensuring a fair assessment.

### A. Need for Model Comparison

There are several reasons why it was necessary to evaluate multiple models in this project:

- Benchmarking Performance Across Diverse Models: Each of the selected models represents different architectures that have shown success in various image classification tasks. By training them on the same dataset with identical hyperparameters, we were able to benchmark their performance under similar conditions and determine which architecture was best suited for our specific task.

  - ResNet models are known for their depth and ability to capture intricate details, which is essential for detecting subtle defects.
  - ResNeXt models offer a balanced trade-off between accuracy and computational efficiency, making them attractive for large-scale applications.
  - EfficientNet models are designed for scaling across multiple dimensions (depth, width, resolution) while maintaining efficient use of computational resources, making them appealing for real-time applications.
  - YOLOv8 and YOLOv11 are specialized in real-time object detection, making them excellent candidates for surgical instrument classification and defect detection due to their speed and accuracy.

- Consistency in Evaluation: To ensure a fair comparison, it was crucial to maintain consistency in hyperparameters such as learning rate, batch size, and training epochs. This way, any performance differences could be attributed to the model architectures themselves, rather than variations in the training process. A consistent hyperparameter setup allowed us to see how each model architecture handled the same data and learned the same defect features, helping us determine the best performing model based on pure architectural capabilities rather than tuning alone.

- Understanding Model Strengths in Specific Contexts: Each model may excel in different contexts. EfficientNet models are known for their ability to handle a broad range of visual features, making them highly generalizable. ResNet-152 is a deeper network, well-suited for identifying subtle or intricate details in images, such as tiny scratches or pores on surgical instruments. ResNeXt-101 offers a unique grouped convolution structure, which can help the model process complex images more efficiently. YOLOv8 and YOLOv11 are designed for real-time detection tasks, making them ideal for high-speed environments where fast decision-making is critical, such as in medical settings.
  By training these models on the same dataset, we were able to observe how each model handled different defect types, which provided insights into their strengths and weaknesses. For example, ResNet-152 may excel at capturing fine details in instruments with small, subtle defects, while YOLOv8 may outperform others in real-time settings where speed is essential.

- Application-Specific Requirements: The surgical instrument defect detection task requires a balance between accuracy and speed. While accuracy is paramount for ensuring the integrity of surgical tools, the system must also provide real-time or near-real-time results to be practical in a clinical setting. Hence, evaluating YOLOv8 and YOLOv11 was essential, as these models are optimized for high-speed detection without compromising accuracy. Comparing these models to EfficientNet-b4, ResNet-152, and ResNeXt-101 allowed us to determine whether a general-purpose model could meet our needs or if specialized real-time models like YOLO were necessary.

- Insight into Generalization and Overfitting: Training multiple models on the same dataset also allowed us to assess how well each model generalizes to new, unseen data. Models like EfficientNet-b4 are known for their scalability and regularization techniques, which help mitigate overfitting. By comparing it with deeper models like ResNet-152 and faster models like YOLOv8, we were able to observe how well each model balanced generalization with performance on the training data. This was especially important for detecting defects across a variety of instruments, lighting conditions, and image quality variations.

## B. Results of the Comparative Analysis

From the analysis, YOLOv8 emerged as the top performer in terms of both accuracy and speed for defect detection across all instrument types. It consistently outperformed the other models in detecting and classifying defects such as corrosion, cuts, scratches, and pores, achieving near-perfect accuracy across all test cases.

ResNet-152 and ResNeXt-101 also performed well, particularly in capturing intricate details in the defect classification stage. However, they required more computational resources and were not as efficient in real-time settings.

EfficientNet-b4, known for its scalability and efficiency, delivered good results, particularly in terms of generalization and handling diverse image conditions, but it did not match YOLOv8's real-time detection capabilities.

The decision to train EfficientNet-b4, ResNet-152, ResNeXt-101, YOLOv8, and YOLOv11 on the same dataset using identical hyperparameters allowed us to make an informed choice for defect classification. By applying consistent training parameters, we ensured that each model was evaluated fairly. The results demonstrated that **YOLOv8** was the best-suited model for this task, offering a perfect balance of accuracy and speed for real-time defect detection in surgical instruments.

The Table V below indicates the comparative analysis results for Bandage scissors.

### TABLE V
### COMPARISON OF BANDAGE SCISSOR INSTRUMENT AND DEFECT CLASSIFICATION PERFORMANCE ACROSS MODELS

| Model | Training Acc. | Testing Acc. | Precision | Recall | F1-Score | ROC-AUC |
|---|---|---|---|---|---|---|
| EfficientNet-b4 | 0.9389 | 0.9907 | 0.9898 | 0.9900 | 0.9899 | 0.9997 |
| ResNet-152 | 0.9375 | 0.9278 | 0.9334 | 0.9291 | 0.9271 | 0.9976 |
| ResNeXt-101 | 0.9539 | 0.9115 | 0.9298 | 0.8976 | 0.8959 | 0.9980 |
| YOLOv8 | **0.9940** | **0.9939** | **0.9936** | **0.9929** | **0.9932** | **0.9999** |
| YOLOv11 | 0.9940 | 0.9907 | 0.9895 | 0.9902 | 0.9897 | 0.9998 |

Bold values indicate the highest metrics achieved by YOLOv8.

The comparative analysis for Dressing Forceps is given in Table VI below.

### TABLE VI
### COMPARISON OF DRESSING FORCEPS INSTRUMENT AND DEFECT CLASSIFICATION PERFORMANCE ACROSS MODELS

| Model | Training Acc. | Testing Acc. | Precision | Recall | F1-Score | ROC-AUC |
|---|---|---|---|---|---|---|
| EfficientNet-b4 | 0.9323 | 0.9849 | 0.9860 | 0.9860 | 0.9859 | 1.0000 |
| ResNet-152 | 0.8999 | 0.8506 | 0.9041 | 0.8530 | 0.8754 | 0.9905 |
| ResNeXt-101 | 0.9326 | 0.9090 | 0.9293 | 0.9159 | 0.9222 | 0.9978 |
| YOLOv8 | **0.9960** | **0.9958** | **0.9983** | **0.9961** | **0.9972** | **1.0000** |
| YOLOv11 | 0.9900 | 0.9898 | 0.9885 | 0.9891 | 0.9888 | 0.9998 |

Bold values indicate the highest metrics achieved by YOLOv8.

Similarly, the comparative analysis for Ex-Probe is given in the Table VII below.

### TABLE VII
### COMPARISON OF EX-PROBE INSTRUMENT AND DEFECT CLASSIFICATION PERFORMANCE ACROSS MODELS

| Model | Training Acc. | Testing Acc. | Precision | Recall | F1-Score | ROC-AUC |
|---|---|---|---|---|---|---|
| EfficientNet-b4 | 0.9384 | 0.9882 | 0.9875 | 0.9911 | 0.9891 | 0.9999 |
| ResNet-152 | 0.9276 | 0.9539 | 0.9543 | 0.9589 | 0.9566 | 0.9704 |
| ResNeXt-101 | 0.9402 | 0.9501 | 0.9506 | 0.9614 | 0.9559 | 0.9984 |
| YOLOv8 | **0.9930** | **0.9934** | **0.9974** | **0.9966** | **0.9970** | **1.0000** |
| YOLOv11 | 0.9770 | 0.9809 | 0.9809 | 0.9782 | 0.9795 | 0.9897 |

Bold values indicate the highest metrics achieved by YOLOv8.

These findings not only validate the robustness of our approach in automated defect classification but also highlight its superior ability to balance high detection accuracy with computational efficiency, making it a scalable solution for industrial applications. The model's real-time inference capabilities, coupled with statistically validated preprocessing techniques, ensure reliable performance across diverse imaging conditions. Furthermore, the significant advantage in inference speed over traditional CNN architectures demonstrates the feasibility of deploying SurgScan in high-throughput production lines where real-time defect detection is crucial. While SurgScan achieves exceptional classification accuracy, the results also indicate areas for future refinement, such as improving detection for low-contrast defects and further optimizing dataset diversity. These insights contribute to the growing need for AI-driven quality control solutions in medical manufacturing, paving the way for enhanced defect detection methodologies that minimize human error and improve surgical instrument reliability.

## X. PROTOTYPE OF AUTOMATED OPTICAL INSPECTION TOOL

The inspection and maintenance of surgical instruments are critical components of quality assurance in healthcare settings, particularly in surgical environments where instrument integrity directly impacts patient safety. Traditional inspection methods rely heavily on visual examination by trained personnel, which, while effective in certain contexts, suffers from limitations including inconsistency, fatigue-induced errors, and an inability to detect subtle or microscopic defects consistently across large volumes of instruments. These limitations are further exacerbated by increasing demands for faster turnaround times in instrument reprocessing workflows, especially in high-volume clinical settings.

While developing robust models and curating a scalable dataset for defect classification are crucial steps, it is equally important to integrate these capabilities into a practical, real-time tool that can be used in real-world applications. In medical settings, particularly during surgical procedures or instrument preparation, any delay in identifying defects on instruments can have serious consequences. Therefore, a tool that can seamlessly classify defects in real-time, while being user-friendly, is necessary for clinical environments.

Automated surgical inspection systems offer a promising alternative by integrating computer vision and machine learning for systematic defect detection. Such systems have been widely used in industrial domains—especially for tasks where precision and repeatability are critical. In the medical domain, however, the deployment of AOI tools for surgical instrument inspection remains an emerging field, with few studies exploring real-time, AI-driven solutions. Campos-Montes demonstrated the feasibility of using convolutional neural networks (CNNs) for classifying surgical instruments based on shape and type, emphasizing the role of high-quality image data and robust models. However, the challenge of defect detection—where features can be subtle and context-dependent—necessitates more advanced object detection architectures capable of localizing and classifying defects simultaneously.

Recent advances in object detection, particularly YOLO (You Only Look Once) architectures, have enabled real-time performance with high accuracy on embedded and resource-constrained devices. The YOLOv8 model, for instance, offers improved inference speed and precision, making it suitable for deployment in practical automated inspection systems for medical applications. These systems not only automate defect identification but also ensure reproducibility, minimize human error, and enable rapid scaling of inspection workflows.

Moreover, studies have shown that integrating these systems in surgical workflows can significantly reduce the incidence of undetected defects, thereby reducing the risk of surgical site infections (SSI) and associated postoperative complications [20]. By leveraging real-time defect detection, healthcare providers can enforce stricter sterilization and maintenance protocols, ultimately enhancing patient safety and operational efficiency.

This prototype represents a step toward addressing these challenges by providing a scalable, AI-powered AOI system tailored for surgical instrument inspection. It builds upon prior research in medical imaging and industrial inspection to deliver a practical solution capable of enhancing quality control in surgical settings.

## XI. System Architecture

The system architecture of the web-based tool for SurgScan is structured in a Client-Server architecture with the following key layers:

- Frontend (Client-Side): Handles the user interface and interaction with users (Admin & User).

- Backend (Server-Side): Manages business logic, database interactions, and image processing.
- Database: Stores persistent user data, batch information, and image details.
- Image Processing Module: Processes images uploaded by users, applying transformations for batch inspections.

The Architecture type of the system is 3-Tier with the Presentation Layer made with React, The Business Logic Layer with Django Backend (REST API), and the Data Layer made up of PostgreSQL.

### A. Hardware Specifications

The hardware specifications for the web-based tool are as follows:

**Server Requirements:**
- Web Server:
  - CPU: 4-16 core processor
  - RAM: 16-32 GB
  - Storage: 500 GB - 1 TB SSD (based on database size and assets)
  - Operating System: Linux (Ubuntu 20.04 LTS or Windows 10/11)
- Database Server:
  - CPU: 4-16 core processor
  - RAM: 16-32 GB
  - Storage: 1 TB SSD (depending on the data volume)
  - Operating System: Linux (Ubuntu 20.04 LTS or Windows 10/11)

### B. Software Stack

The Software stack used in SurgScan is as follows:

- Frontend:
  - React.js: For rendering components, handling routing (react-router-dom), and managing user interactions
  - CSS: For layout, design, and responsive styling.
- Backend:
  - Django (Python): REST API to manage user authentication, batch data, and image processing requests.
  - Django Rest Framework (DRF): For building RESTful API endpoints.
  - Image Processing: OpenCV and Pillow libraries for image manipulation (sharpening, cropping, resizing).
- Database:
  - PostgreSQL: Stores user information, batch details, image metadata, and inspection results.
- File Storage:
  - Django Default Storage: For handling file uploads, including user profile images and batch inspection images.
- Authentication:
  - Django Authentication: Manages user login and roles (Admin, User).

## C. Core Components and Modules

The core components of SurgScan are as follows: **Front end:**

It consists of User interfaces, routing, form handling, and UI components.

- User Interfaces:
  - Admin Dashboard: Allows administrators to manage users, batches, and view statistics.
  - User Dashboard: For regular users to upload images, view batch details, and start inspections.
- Form Handling:
  - Users can submit forms (image upload, registration, etc.) and interact with backend APIs for data submission.
- Routing:
  - react-router-dom is used for navigation between different views (e.g., user dashboard, admin dashboard).
- UI Components:
  - Batch Table: Displays batch details such as batch number, total images inspected, and defect analysis.
  - Admin User Management: Allows activating or deactivating users.

**Backend (Django):**

- API Layer:
  - REST API built using Django Rest Framework (DRF) for user and batch management, image processing, and authentication.
- Views:
  - Admin Views: Views for handling user management, batch retrieval, and statistics.
  - User Views: Views for image uploading, inspection management, and batch creation.
- Image Processing Pipeline:
  - OpenCV and Pillow libraries are used for image manipulation (resize, unsharp masking, cropping). Steps include reading images, applying transformations, resizing while maintaining aspect ratio, and preparing them for inspection.
- Business Logic:
  - User Management: Admins can deactivate users by changing their status to "inactive."
  - Batch Processing: System checks if a user is already assigned to a batch before allowing image uploads.

**Database (PostgreSQL)**

- User table:
  - Stores user information such as email, password (hashed), profile image URL, status (active/inactive), and roles (Admin, User).
- Batch Table:
  - Stores batch-related information including the Batch number, Total images inspected, defected/non-defected images and image metadata.

## XII. EVALUATION METRICS

The evaluation metrics for the testing we performed using Postman (for API testing) and Django unit testing would primarily revolve around the functionality, correctness, performance, and security of the backend.

**API Testing Using Postman:** API testing focuses on ensuring that the Django backend APIs work as intended, respond with the correct data, and handle edge cases appropriately. The key metrics for API testing are:

- Functionality:
  - Correctness of Responses: Ensuring that the APIs return the expected data for valid requests (e.g., correct defect classification, correct instrument identification).
  - HTTP Status Codes: Verifying that the API returns appropriate HTTP status codes for each scenario (e.g., 200 for success, 400 for bad requests, 404 for resource not found, 500 for server errors).
  - Data Validation: Ensuring that the backend correctly validates input data (e.g., missing fields, invalid image formats, invalid batch numbers) and returns meaningful error messages.
- Security:
  - Authentication and Authorization: If your APIs are protected (e.g., login API, batch creation API), testing whether unauthorized users are blocked from accessing restricted resources.
  - Data Protection: Ensuring that sensitive data (like user credentials) is not exposed in API responses and is securely transmitted.
- Error handling:
  - Graceful Error Responses: Verifying that the API handles invalid input gracefully without crashing and provides meaningful error messages (e.g., misclassified defects due to poor image quality should return informative error responses).
- Key Metrics for API Testing:
  - Success Rate: The percentage of API requests that are correctly handled (i.e., returning a 200 status code and correct data).
  - Error Rate: The percentage of API requests that fail (return a 4xx or 5xx status code).

**Unit Testing Using Django:** Django unit tests typically focus on testing individual components of the backend, including models, views, and forms, ensuring they function correctly in isolation. The evaluation metrics for unit testing would include:

- Test Coverage:
  - Code Coverage: The percentage of code that is executed by the unit tests. High code coverage ensures that most of your code is being tested.
- Functionality:
  - Correctness of Models: Verifying that the Django models behave as expected (e.g., user creation, batch detail creation, image processing logic).

– View Responses: Ensuring that the views (API end-points) return the correct HTTP responses, with the expected data and status codes.

– Forms and Validation: Testing that Django forms correctly validate input data and reject invalid submissions (e.g., checking user registration or image upload forms).

• Edge Case Handling:

– Boundary Testing: Testing how the system handles edge cases (e.g., very large images, incorrect file formats, missing data, invalid credentials).

– Error Handling: Ensuring that the system raises appropriate exceptions and returns meaningful error messages (e.g., handling non-existent batches or defective image uploads).

• Key Metrics for Unit Testing:

– Test Case Pass Rate: The percentage of test cases that pass without errors. A high pass rate indicates the functionality is working correctly.

– Code Coverage: The percentage of code executed by the tests, ensuring that all critical paths are tested.

## XIII. Overview of Tool

Recognizing the need for an integrated solution, we developed a web application that combines instrument classification and defect detection in a single, user-friendly interface. This tool serves as the final layer in the system, where all the backend models (such as YOLOv8 for instrument and defect classification) and the curated dataset are put into practical use.

### A. Data Flow and System Interactions

• Admin Workflows:

– Login: Admin logs in via a form that interacts with the Django backend.

– Manage Users: Admin can view a list of users (active/inactive) and deactivate users through the API.

– Manage Batches: Admins can view batch details and associated inspection results, along with statistics.

– View Stats: Admins have a dashboard displaying batch statistics and graphs generated in React.

• User Workflows:

– Login: Users log in and access their dashboard.

– Upload Images: Users upload images for inspection. These images are processed using the image pipeline and saved in storage.

– Batch Management: Users are assigned to batches, and the system ensures batch integrity before image submission.

– View Inspection Results: Users can see inspection results (defected, non-defected images) for each batch.

• Image Processing Workflow:

– Upload: User uploads an image file through the web form.

– Processing: The backend applies image transformations (resize, crop, enhance) using OpenCV and Pillow.

– Save: The processed image is saved, and metadata is stored in the database.

– Batch Assignment: The image is linked to the respective batch, and inspection results are updated.

• Security Considerations:

– User Authentication: Managed through Django's built-in authentication (optional JWT tokens can be added for API security).

• Role-Based Access Control (RBAC): Admins and Users have different access rights. Admins manage users and batches, while users can only upload and inspect images.

• Data Validation: The backend validates all inputs from the frontend (e.g., batch creation, image uploads) to ensure data integrity.

The tool named "SurgScan" logs in the user using the credentials and takes to the home screen where the classification of isntrument or defect can be done as shown in Figure 31.

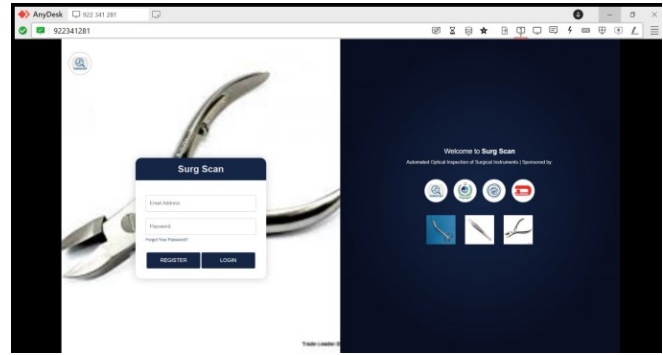

Fig. 31. Overview of the Login page of SurgScan

Furthermore, the overall stats of the batch can also be seen that details the number of defected and undefected surgical instruments in that batch as shown in Figure 32.

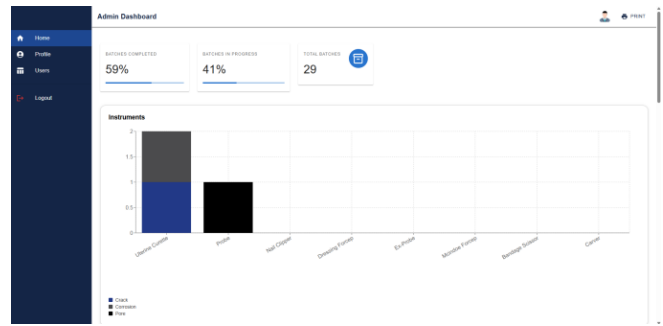

Fig. 32. Overview of the Statistics page of SurgScan

Moreover, the dashboard also displays the admin and users of the system as shown in Figure 33.

They can be edited/deleted/added as per admin preference or usage. This tool is particularly valuable in settings where large volumes of instruments need to be inspected quickly and accurately. Sterilization units in hospitals, where instruments are cleaned and prepared for surgeries, can use this web application to automatically screen instruments for defects, reducing the workload on staff and improving overall efficiency. In addition, it allows medical staff to focus on more critical tasks, knowing that the integrity of the instruments has been reliably checked. This real-time defect classification tool ensures that the models we developed for defect detection can be directly applied in a clinical environment, making surgical procedures safer and more reliable.

Fig. 33. Overview of SurgScan Admin Dashboard

## Acknowledgment

This work is supported by the National Research Program for Universities (NRPU), Higher Education Commission (HEC) Pakistan, under Project No. 15872, titled "Automated Optical Inspection for Industrial Surgical Instruments." Additonally, we would like to Acknowledge the Industrial partners Dr. Frigz International and Daddy D Pro for their industrial collaboration and expert guidance for the completion of the project.

## REFERENCES

[1] W. R. Waked, A. K. Simpson, C. P. Miller, D. P. Magit, and J. N. Grauer, "Sterilization wrap inspections do not adequately evaluate instrument sterility," *Clin. Orthop. Relat. Res.*, vol. 462, pp. 207–211, Sep. 2007, doi: 10.1097/BLO.0b013e318065b0bc.

[2] S. A. Singh and K. A. Desai, "Automated surface defect detection framework using machine vision and convolutional neural networks," *J. Intell. Manuf.*, vol. 34, no. 4, pp. 1995–2011, 2023, doi: 10.1007/s10845-022-02064-6.

[3] A. Singh and P. Desai, "Deep learning-based surface defect detection in additive manufacturing," *J. Intell. Manuf.*, vol. 34, no. 8, pp. 3456–3478, 2023, doi: 10.1007/s10845-023-02191-5.

[4] A. Fis¸ne *et al.*, "Lightweight deep learning models for real-time defect detection," *J. Intell. Manuf.*, vol. 35, pp. 1123–1145, 2024, doi: 10.1007/s10845-024-02509-3.

[5] D. Tabernik, S. Šela, J. Skvarč, and D. Skočaj, "Segmentation-based deep-learning approach for surface-defect detection," *J. Intell. Manuf.*, vol. 31, no. 3, pp. 759–776, 2020, doi: 10.1007/s10845-019-01476-x.

[6] C.-M. Lo and T.-Y. Lin, "Automated optical inspection based on synthetic mechanisms combining deep learning and machine learning," *J. Intell. Manuf.*, vol. 36, no. 7, pp. 4769–4783, 2025, doi: 10.1007/s10845-025-02345-8.

[7] F. El Kalach, I. Yousif, T. Wuest, A. Sheth, and R. Harik, "Cognitive manufacturing: definition and current trends," *J. Intell. Manuf.*, vol. 36, pp. 3695–3715, 2025.

[8] H. Zhang, S. D. Semujju, Z. Wang *et al.*, "Large scale foundation models for intelligent manufacturing applications: a survey," *J. Intell. Manuf.*, 2025.

[9] Y. Wang, W. Xu, C. Wang, Y. Huang, and Z. Zhang, "Time series fault prediction via dual enhancement," *J. Intell. Manuf.*, vol. 36, pp. 5247–5262, 2025.

[10] M. Javaid, A. Haleem, R. P. Singh, and R. Suman, "Industry 5.0: Potential applications in COVID-19," *J. Ind. Integr. Manage.*, vol. 8, no. 01, pp. 17–41, 2023.

[11] X. Xu, Y. Lu, B. Vogel-Heuser, and L. Wang, "Human-centric AI for Industry 5.0: A review of key enabling technologies," *IEEE Trans. Ind. Informat.*, vol. 19, no. 3, pp. 2315–2325, 2023.

[12] J. Wang, Y. Li, and H. Zhang, "Deep learning-based surface defect detection for highly reflective metal parts: A survey," *Robot. Comput.-Integr. Manuf.*, vol. 85, Art. no. 102636, 2024.

[13] V. V. Bhandarkar *et al.*, "An overview of traditional and advanced methods to detect part defects in additive manufacturing processes," *J. Intell. Manuf.*, vol. 36, no. 7, pp. 4411–4446, 2025, doi: 10.1007/s10845-025-02312-4.

[14] M. B. Abbas, A. S. Prabuwono, S. N. H. S. Abdullah, and R. I. Hussain, "Automated visual inspection for surgical instruments based on spatial feature and K-Nearest neighbor algorithm," *Adv. Sci. Lett.*, vol. 20, no. 1, pp. 153–157, 2014, doi: 10.1166/asl.2014.1178.

[15] Y. Zhi *et al.*, "End-to-end deep learning framework for welding defect detection," *J. Intell. Manuf.*, vol. 35, pp. 789–805, 2024, doi: 10.1007/s10845-024-02314-y.

[16] D. O. Pontes *et al.*, "Adenosine triphosphate (ATP) sampling algorithm for monitoring the cleanliness of surgical instruments," *PloS One*, vol. 18, no. 8, Art. no. e0284967, 2023.

[17] B. Ran, B. Huang, S. Liang, and Y. Hou, "Surgical instrument detection algorithm based on improved YOLOv7x," *Sensors*, vol. 23, no. 11, Art. no. 5037, 2023.

[18] S. Bodenstedt *et al.*, "Real-time image-based instrument classification for laparoscopic surgery," 2018. [Online]. Available: https://arxiv.org/abs/1808.00178

[19] K. Lam *et al.*, "Deep learning for instrument detection and assessment of operative skill in surgical videos," *IEEE Trans. Med. Robot. Bionics*, vol. 4, no. 4, pp. 1068–1071, 2022.

[20] A. Zia *et al.*, "Surgical tool classification and localization: results and methods from the miccai 2022 surgtoolloc challenge," 2023. [Online]. Available: https://arxiv.org/abs/2305.07152

[21] A. J. M. Plompen *et al.*, "The joint evaluated fission and fusion nuclear data library, JEFF-3.3," *Eur. Phys. J. A*, vol. 56, pp. 1–108, 2020.

[22] G. Jocher, A. Chaurasia, and J. Qiu, "YOLO by Ultralytics," 2023. [Online]. Available: https://github.com/ultralytics/ultralytics

[23] F. A. Ahmed *et al.*, "Deep learning for surgical instrument recognition and segmentation in robotic-assisted surgeries: a systematic review," *Artif. Intell. Rev.*, vol. 58, no. 1, Art. no. 1, 2024.

[24] E. W. Stockert and A. Langerman, "Assessing the magnitude and costs of intraoperative inefficiencies attributable to surgical instrument trays," *J. Amer. Coll. Surg.*, vol. 219, no. 4, pp. 646–655, 2014.

[25] J. M. Mhlaba, E. W. Stockert, M. Coronel, and A. J. Langerman, "Surgical instrumentation: the true cost of instrument trays and a potential strategy for optimization," *J. Hosp. Admin.*, vol. 4, no. 6, pp. 82–88, 2015.

[26] A. O. Campos-Montes, C. J. Vega-Urquizo, M. G. S. P. Paredes, and K. Acuna-Condori, "Implementation of a computer-assisted surgical instrument sterilization system based on deep learning for health centers," in *Proc. 2022 IEEE ANDESCON*, 2022, pp. 1–6, doi: 10.1109/CASE49997.2022.9926543.

[27] X.-H. Liu, C.-H. Hsieh, J.-D. Lee, S.-T. Lee, and C.-T. Wu, "A vision-based surgical instruments classification system," in *Proc. 2014 Int. Conf. Adv. Robot. Intell. Syst. (ARIS)*, 2014, pp. 72–77.

[28] H.-B. Le *et al.*, "Robust surgical tool detection in laparoscopic surgery using yolov8 model," in *Proc. 2023 Int. Conf. Syst. Sci. Eng. (ICSSE)*, 2023, pp. 537–542.

[29] J. Yu *et al.*, "Performance Evaluation of Surgical Instrument Recognition Based on a series of YOLO Models," in *Proc. 2024 9th Int. Conf. Image, Vis. Comput. (ICIVC)*, 2024, pp. 52–58.

[30] McKinsey & Company, "The hidden cost of quality in medical devices: Reducing scrap, rework, and rejection through intelligent automation," Healthcare Systems & Services Practice, Tech. Rep., 2023. [Online]. Available: https://www.mckinsey.com/industries/life-sciences/our-insights/reducing-quality-costs-in-medical-device-manufacturing

[31] J. E. See, "Visual inspection: a review of the literature," Sandia National Laboratories, Albuquerque, NM, USA, Tech. Rep. SAND2012-8590, 2012.

[32] Steelco Group, "SUIS – Surgical Instrument Vision System," 2025. [Online]. Available: https://www.steelcogroup.com/us/technology/surgical-instrument-scanner/

[33] Daddy D Pro, "Surgical and Dental Instruments," 2025. [Online]. Available: https://daddydpro.com/index.php

[34] American College of Surgeons, "What are the surgical specialties?" 2021. [Online]. Available: https://www.facs.org/education/resources/medical-students/faq/specialties

[35] USMS, "New Trends in Surgical Instruments," 2018. [Online]. Available: https://www.usms.biz/trends-surgical-instruments/

[36] National University of Singapore, "Intelligent Surgical Tools Inspection System," 2021. [Online]. Available: https://cde.nus.edu.sg/edic/projects/innovating-for-better-healthcare/surgical-tools-inspection/

[37] FACS, "Commonly Used Surgical Instruments," 2021. [Online]. Available: https://www.facs.org/media/wgcmalet/commonsurgicalinstrumentsmodule.pdf

[38] Pakistan Business Council, "Enhancing the Competitiveness of Pakistan's Surgical Instruments Industry," 2021. [Online]. Available: https://www.pbc.org.pk/research/enhancing-the-competitiveness-of-pakistans-surgical-instruments-industry/

[39] TRTA, "Surgical Instruments," 2021. [Online]. Available: http://trtapakistan.org/sector-products/industrial-products/surgical-instruments/

[40] Ministry of Finance, Pakistan, "Trade and Payments," 2021. [Online]. Available: https://www.finance.gov.pk/survey/chapter22/PES08-TRADE.pdf